%% file: full-paper.tex
\documentclass{book}    

\usepackage{piers}  
\usepackage[]{graphicx}
\graphicspath{{Figures/}}
\usepackage{graphicx}

\usepackage{amsmath}
\usepackage{amssymb}
\usepackage{epsfig}
\usepackage{cite}
\usepackage{color}
\usepackage{balance}
\usepackage{pgfplots}
\usepackage{wrapfig}
\usepackage{lipsum}
\usepackage{hyperref}
\usepackage{gensymb}
\usepackage{bm}
\usepackage{breqn}
\usepackage{enumitem}
\usepackage{tabto}

\begin{document}

\title{The Effectiveness of Edge Detection Evaluation Metrics for Automated Coastline Detection}
\maketitle

\author      {F. M. Lastname}
\affiliation {University}
\address     {}
\city        {Boston}
\postalcode  {}
\country     {USA}
\phone       {345566}    
\fax         {233445}    
\email       {email@email.com}  
\misc        { }  
\nomakeauthor

\author      {F. M. Lastname}
\affiliation {University}
\address     {}
\city        {Boston}
\postalcode  {}
\country     {USA}
\phone       {345566}    
\fax         {233445}    
\email       {email@email.com}  
\misc        { }  
\nomakeauthor

\begin{authors}

{\bf Conor~O'Sullivan}$^{1,2}$ $^{\dagger}$, 
{\bf Seamus~Coveney}$^{3}$, 
{\bf Xavier~Monteys}$^{4}$, 
{\bf and Soumyabrata Dev}$^{1,2}$\\
\medskip

$^{1}$ADAPT SFI Research Centre, Dublin, Ireland\\

$^{2}$School of Computer Science, University College Dublin, Ireland \\

$^{3}$Envo-Geo Environmental Geoinformatics, Skibbereen, Ireland \\

$^{4}$Geological Survey Ireland, Dublin, Ireland

$^{\dagger}$ Presenting author and corresponding author

\end{authors}

\begin{paper}

\begin{piersabstract}
We analyse the effectiveness of RMSE, PSNR, SSIM and FOM for evaluating edge detection algorithms used for automated coastline detection. Typically, the accuracy of detected coastlines is assessed visually. This can be impractical on a large scale leading to the need for objective evaluation metrics. Hence, we conduct an experiment to find reliable metrics. We apply Canny edge detection to 95 coastline satellite images across 49 testing locations. We vary the Hysteresis thresholds and compare metric values to a visual analysis of detected edges. We found that FOM was the most reliable metric for selecting the best threshold. It could select a better threshold 92.6\% of the time and the best threshold 66.3\% of the time. This is compared RMSE, PSNR and SSIM  which could select the best threshold 6.3\%, 6.3\%  and 11.6\% of the time respectively. We provide a reason for these results by reformulating RMSE, PSNR and SSIM in terms of confusion matrix measures. This suggests these metrics not only fail for this experiment but are not useful for evaluating edge detection in general. 

\end{piersabstract}

\psection{Introduction}
Monitoring the coastline is important for coastal management and being prepared for natural disasters. Yet, manually identifying and mapping the coastline can be a time-consuming process. Developing methods for automatically detecting the coastline can improve the efficiency and accuracy of coastal monitoring efforts. Following this, it is crucial to develop metrics for assessing the accuracy of these automated methods. They will allow researchers and practitioners to objectively evaluate automated coastline detection methods and compare their performance across different datasets and conditions. This comparison can also be done on a scale that is not practical with visual analysis. 

\psection{Related Work}
Edge detection is an important aspect of coastal monitoring processes. We can detect the border where land meets water~\cite{vukadinov2017algorithm, klinger2011antarctic, paravolidakis2018automatic, seale2022swed} or the vegetation line along the coast~\cite{rogers2021vedge_detector}. These lines can be considered prominent edges that we aim to detect using automated edge detection methods. Detecting them is the first step in more complex monitoring processes. For example, by tracking the lines through time we can estimate coastal erosion rates~\cite{luijendijk2018state}. 

The most common way for evaluating the accuracy of the edge detection methods is using visual analysis~\cite{vukadinov2017algorithm, klinger2011antarctic, paravolidakis2018automatic}. This involves comparing the images of the coastlines to the detected edges and making a subjective evaluation of the performance.  For these studies, visual analysis is an appropriate method. The datasets are small and they lack ground truth edges for which to compare the detected edges too. 

More objective evaluation metrics have been applied in studies with larger datasets~\cite{seale2022swed}. Here the researchers compared different image segmentation models on a dataset of 98 coastline satellite images. The researchers took a segmentation approach to coastline detection. This means the output of the model was a binary segmentation mask that classified each pixel as either land (0) or water (1). The output was then compared to ground truth masks using metrics such as accuracy, precision and recall. It is not clear if these metrics are appropriate for evaluating edge detection algorithms.  

This would involve comparing predicted edges to ground truth edges. Various quality assessment metrics have been proposed to make this comparison. These include root mean square error (RMSE), peak signal-to-noise ratio (PSNR), universal image quality index (UQI), structural similarity (SSIM)~\cite{sadykova2017quality,tariq2021quality} and figure of merit (FOM)~\cite{tariq2021quality}. We aim to understand how effective these metrics are for evaluating coastline edge detection methods.

\psection{Contributions}
 Our main contributions are to:
\vspace{-0.2cm}
\begin{itemize}
    \item Analyse the effectiveness of various edge detection algorithms on a diverse dataset of coastline satellite images. 
    \vspace{-0.2cm}
    \item Reformulate RMSE, PSNR, SSIM in terms of confusion matrix measures to understand why they are not effective metrics for evaluating edge detection algorithms. 
    \vspace{-0.2cm}
    \item We open source the code to reproduce the results of this paper. This can be found in the GitHub repository~\footnote{In the spirit of reproducible research, the codes to reproduce the results of this paper can be found here: \url{https://github.com/conorosully/edge-detection-metrics}.}.
\end{itemize}

\psection{Methodology}
\psubsection{Dataset}

We base our analysis on the Sentinel-2 Water Edges Dataset (SWED)~\cite{seale2022swed}. This contains satellite images and binary segmentation masks from a variety of coastlines. We consider the 98 test images and labels from SWED as a reference dataset.  We do not consider the training set as it has been labelled using semi-supervised methods. In comparison, more care has been taken when labelling the test set. Although, we did identify 3 images in the test set with erroneous binary masks (see appendix for details). We removed these leaving us with 95 images. 

As input, 12 spectral bands are available from sentinel-2 imagery. They include visible red, green and blue wavelengths. The example image in Figure~\ref{fig:dataset} has been created by combining these three bands. The bands also include shortwave inferred wavelengths. We considered each of the bands separately and treat them as grayscale images. 

As seen in Figure~\ref{fig:dataset} the dataset contains a mask for each image. The pixels in the mask are given a value of 0 for land and 1 for water. We obtain edge maps by applying the Canny edge detection algorithm discussed in the next section to the masks. We take these edge maps as our ground truth.

\begin{figure}[h]
\centering
\includegraphics[width=0.99\textwidth]{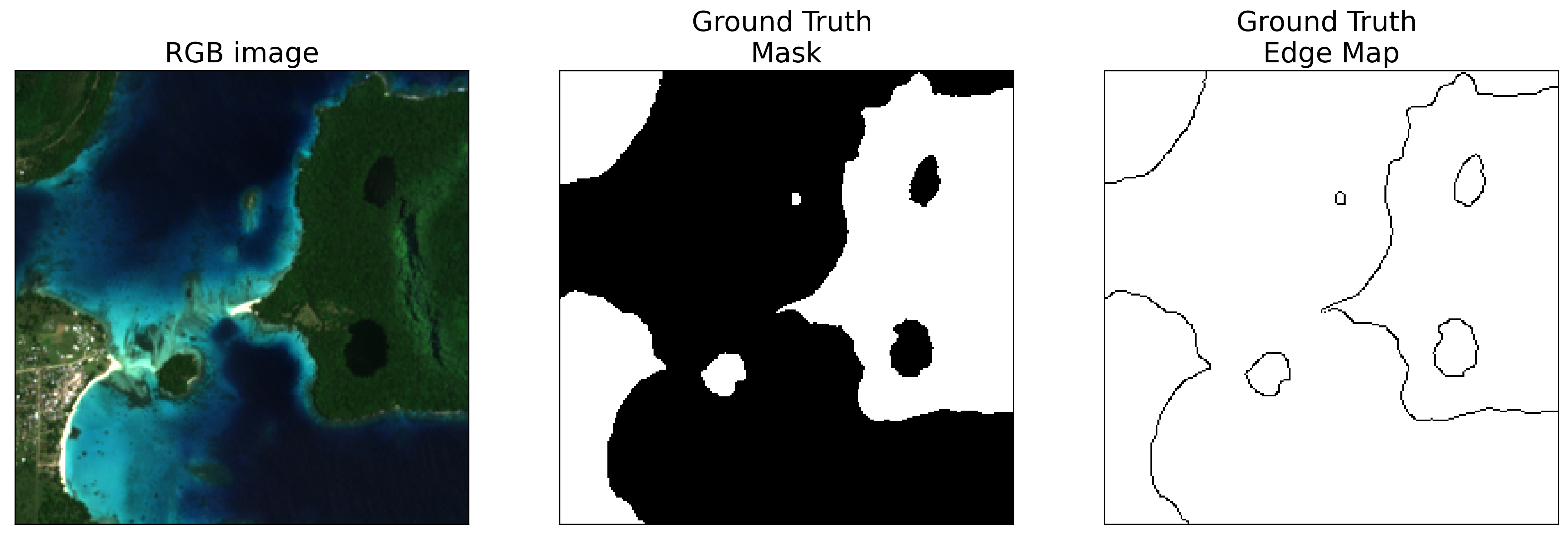}
\caption{Input images and ground truth masks obtained from SWED~\cite{seale2022swed}. The ground truth edge map has been obtained by applying canny edge detection to the ground truth mask.}
\label{fig:dataset}
\vspace{-0.5cm}
\end{figure}

\psubsection{Edge detection}
The Canny edge detection algorithm is a multistage algorithm used to identify prominent edges in images~\cite{canny1986computational}. The steps include:
\begin{enumerate}
    \item Noise reduction using Gaussian smoothing
    \item Calculating gradients by considering changes in pixel intensity
    \item Non-maximum suppression to produce thin edges
    \item Hysteresis thresholding to select the final edges
\end{enumerate}
The final step is of particular importance to this paper. Two thresholds are used. Pixels with an intensity above the upper threshold are considered strong edges. Pixels below the upper threshold but above the lower threshold are considered weak edges. Finally, all strong edges and weak edges connected to strong edges are labelled as edges. 

We apply Canny using different Hysteresis thresholds---[50,100], [50,150], [100,200], [100,300], [200,400], [200,600] with the lower and upper bounds given respectively. By varying the thresholds, we change the sensitivity of the algorithm. Ideally, as we increase the thresholds the algorithm should become less sensitive to noisy edges caused by factors like ocean swell and land development. Then, at some point, even  prominent coastal edges should be excluded. We then compare the detected edges to the ground truth using a variety of metrics. 

\psubsection{Evaluation metrics}
\vspace{0.1cm}
\textbf{Route Mean Square Error} (RMSE) is defined as:

$$ RMSE(E,G) = \sqrt{MSE(E,G)} $$
$$ MSE(E,G) = \frac{1}{MN}  \sum_{i=0}^{M-1} \sum_{j=0}^{N-1} (E_{i,j} - G_{i,j})^2 $$

$E_{i,j}$ and $G_{i,j}$ are the pixel values in the $i,jth$ position in the detected edge map (E) and the ground truth (G). In general, both images will consist of MxN pixels. In our case, both M and N are equal to 256. 

\vspace{0.1cm}

\textbf{Peak Signal-to-noise Ratio} (PSNR) is defined as:

$$ PSNR(E,G) = 10log_{10}\frac{255^2}{MSE(E,G)} $$

Smaller values for RMSE indicate better performance. In comparison, larger values for PSNR indicate better performance. As both metrics are based on MSE, we should expect them to identify similar Hysteresis thresholds as the best performers. 

\vspace{0.1cm}

\textbf{Structural Similarity Index} (SSIM) is defined as: 

$$ SSIM(G, E) = [l(G, E)]^\alpha.[c(G, E)]^\beta.[s(G, E)]^\gamma $$
$$ l(G,E)={\frac {2\mu _{G}\mu _{E}+c_{1}}{\mu _{G}^{2}+\mu _{E}^{2}+c_{1}}} $$
$$ c(G,E)={\frac {2\sigma _{G}\sigma _{E}+c_{2}}{\sigma _{G}^{2}+\sigma _{E}^{2}+c_{2}}} $$
 $$ s(G,E)={\frac {\sigma _{G,E}+c_{3}}{\sigma _{G}\sigma _{E}+c_{3}}} $$

$\mu _{G}$ and $\mu _{E}$ are the average pixel values, $\sigma _{G}$ and $\sigma _{E}$ are the standard deviations of the image pixel values and $\sigma _{G,E}$ is the covariance of the pixel values. SSIM will fall between -1 and 1 with 1 indicating that the images are perfectly identical. RMSE and PSNR both considered pixels on an individual basis. Whereas, SSIM attempts to base the similarity measure on the luminance (l), contrast (c), and structure (s) of the two images. The result is a metric that is more similar to the human visual system for identifying symmetry~\cite{ssimpaper}.

\vspace{0.1cm}

\textbf{Figure of merit} (FOM) is defined as: 
$$ FOM(E,G) = \frac{1}{max(N_E, N_G)} \sum_{k=1}^{N_E} \frac{1}{1+\alpha d^2(k)} $$

$N_G$ is the number of the actual edge pixels, $N_E$ is the number of the detected edge pixels, $\alpha$ is the scaling constant, and $d(k)$ is the minimum distance between the detected edge pixel and an actual edge pixel~\cite{tariq2021quality}.

\vspace{0.1cm}

We also consider \textbf{Confusion Matrix-based Measures} based on the values in Table~\ref{tab:confusion_matrix}. TP is the count of cases where $E_{i,j} = G_{i,j} = 1$, TN is the count where $E_{i,j} = G_{i,j} = 0$, FP is the count where $E_{i,j} = 1, G_{i,j} = 0 $ and FN is the count where $E_{i,j} = 0, G_{i,j} = 1$. We do not evaluate the Canny edge detection algorithm directly with these measures but we considered them when reasoning about the effectiveness of the other metrics. 

\vspace{0.1cm}

\input{figures/confusion_matrix}

Other well-known metrics have been excluded from this study. This includes UQI as it is a special case of SSIM where the stabilizing constants are set to 0~\cite{ssimpaper}. We also excluded the Edge-based structural similarity measure (ESSIM) as it was developed for comparing blurry images ~\cite{sadykova2017quality}.

\psubsection{Visual analysis}
We evaluate the effectiveness of the above metrics to select the best Hysteresis threshold using visual analysis. For each of the 95 test images, we select the threshold that produces the edge map that is visually the most similar to the ground truth edge map. This is done without looking at the evaluation metric values. We can then compare these to the threshold selected by the evaluation metrics. This allows us to calculate the percentages of images correctly identified by each metric. We restrict the visual analysis to the NIR band as it has been shown to be effective for extracting water bodies~\cite{mondejar2019near}. 

\psection{Results and Discussion}

\psubsection{Metric trends by threshold}
Figure~\ref{fig:all_metric} shows the trends in RMSE, PSNR, SSIM and FOM as we increase the thresholds. For all 12 spectral bands, RMSE, PSNR and SSIM indicate a monotonic improvement as the thresholds are increased and the highest threshold produces the best results. In addition, compared to the other bands, these metrics indicate that NIR had the worse average performance at all threshold levels. In comparison, the best threshold for each band varied when using FOM with [100,200], [100,300] being the most common. The NIR band is also not the worse overall performing band. 

\begin{figure}[h]
\centering
\includegraphics[width=0.99\textwidth]{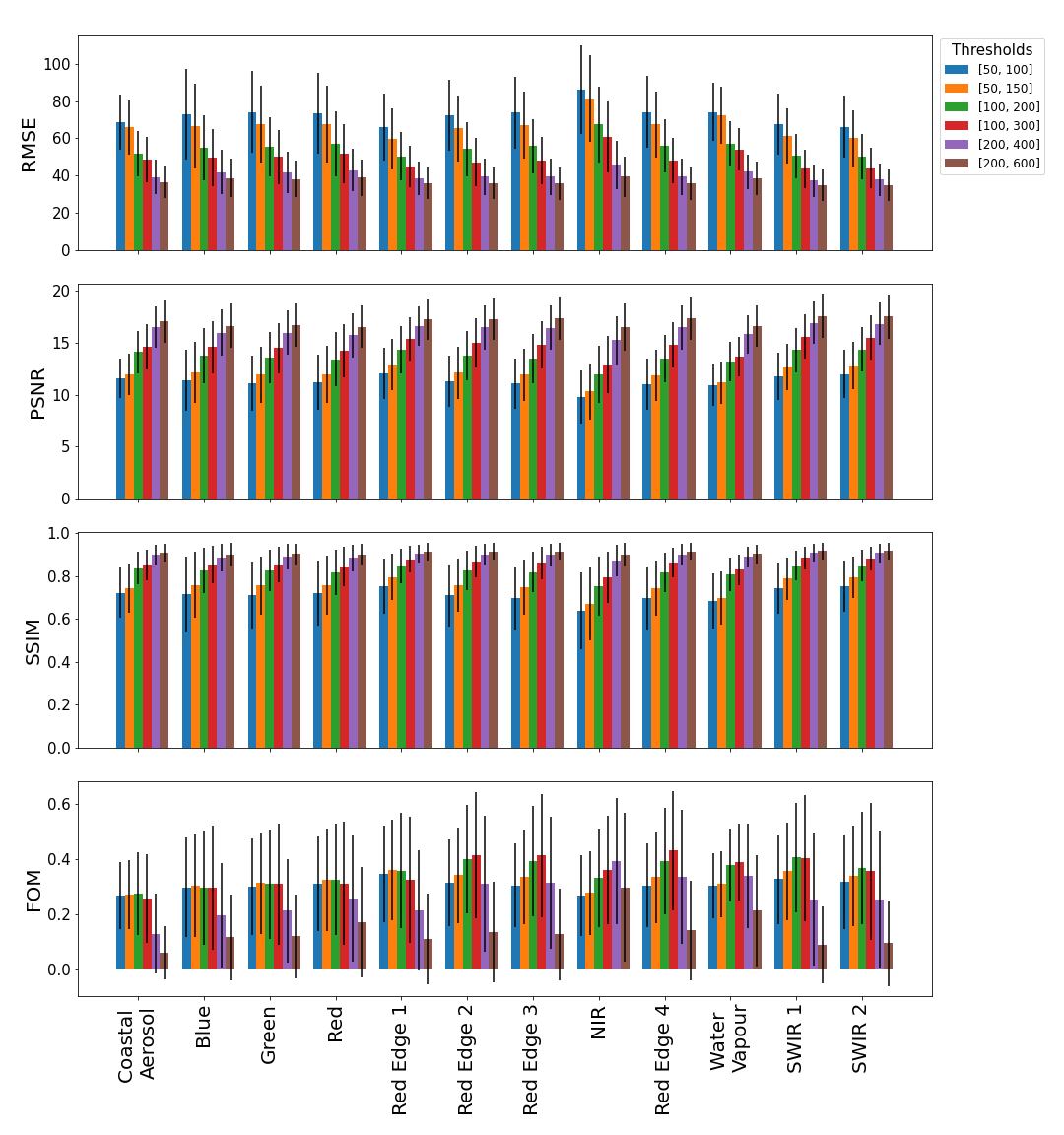}
\caption{Average RMSE, PSNR, SSIM and FOM for each spectral band with error bars given by the standard deviation. Based on RMSE, PSNR and SSIM, the performance improves as we increase the thresholds and the NIR band is the worst-performing band. For FOM, we do not see a monotonic improvement for any of the spectral bands. }
\label{fig:all_metric}
\vspace{-0.5cm}
\end{figure}

Figure~\ref{fig:all_metric} shows that FOM had a larger standard deviation when compared to the other metrics. A potential reason for this is that the best threshold for each image does vary and FOM has a high standard deviation as it is able to identify the most appropriate threshold. We can see this in Table~\ref{tab:threshold_counts}. This gives the counts of test images if we selected the best-performing threshold across all 12 spectral bands. For RMSE, PSNR and SSIM, the counts are skewed towards the higher thresholds. For FOM, the counts are more evenly distributed. 

\input{figures/threshold_counts.tex}
\psubsection{Visual analysis}

Ultimately, the monotonic improvement for RMSE, PSNR and SSIM is suspicious.  It suggests that as we decrease the sensitivity of the Canny algorithm we get better results. Additionally, the poor performance of the NIR bands indicated by the metrics is contrary to prior research. To confirm that FOM is the only metric able to identify appropriate thresholds we performed a visual analysis. This showed that FOM could select the best threshold for 66.3\% of the 95 test images. In comparison, RMSE, PSNR and SSIM selected the best image 6.3\%, 6.3\%  and 11.6\% of the time respectively. 

We can see some examples of this analysis in Figure~\ref{fig:good_examples}. Here we compare the reference edge maps of 4 test images to the predicted edge map using the NIR band at the varying threshold levels. You can see that as we increase the thresholds fewer noisy edges are detected and then later the algorithm detects fewer coastline edges. In each case, FOM selected the edge map that is visually the most similar to the ground truth. In comparison, RMSE, PSNR, and SSIM tend to select the maps with fewer edges. 

\begin{figure}[h]
\centering
\includegraphics[width=0.99\textwidth]{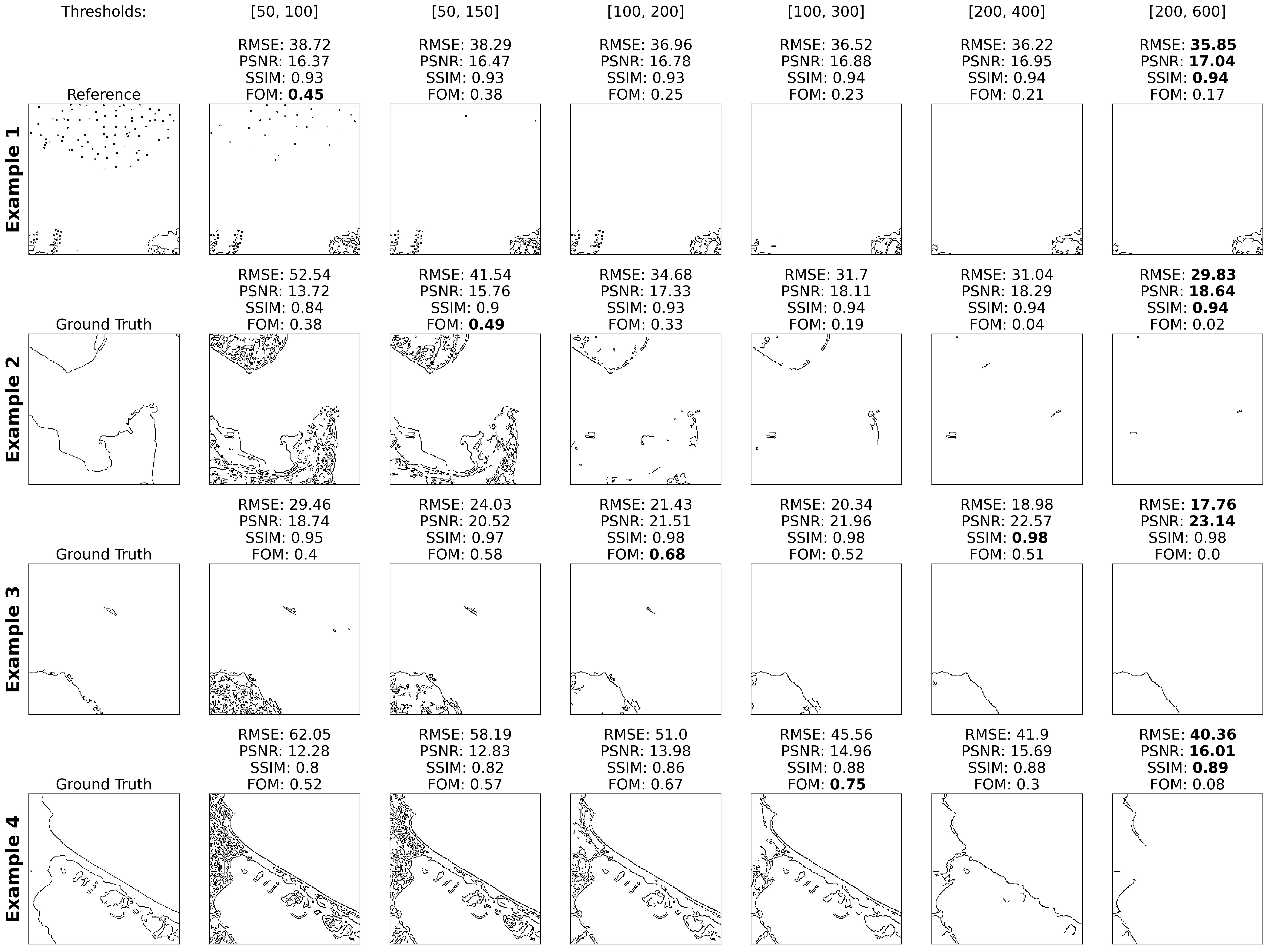}
\caption{Examples of where FOM correctly identifies the detected edge map that is visually the most similar to the ground truth. Each row gives a different example image. The columns give the edge maps detected using the increasing Hysterious thresholds. In each case, the NIR band is used as the input to Canny. The bold numbers give the best value for each metric.}
\label{fig:good_examples}
\vspace{-0.3cm}
\end{figure}

In Figure~\ref{fig:bad_examples}, Example 5 shows an example of where FOM does not clearly select the best edge map. Based on the visual analysis, the threshold of [100,200] was selected whereas FOM selects a threshold of [100,300]. However, the map selected by FOM is visually more similar to the ground truth than the one selected by the other metrics. If we consider all these cases, we found that FOM could select the same or better edge map for 92.6\% of the images.

\begin{figure}[h]
\centering
\includegraphics[width=0.99\textwidth]{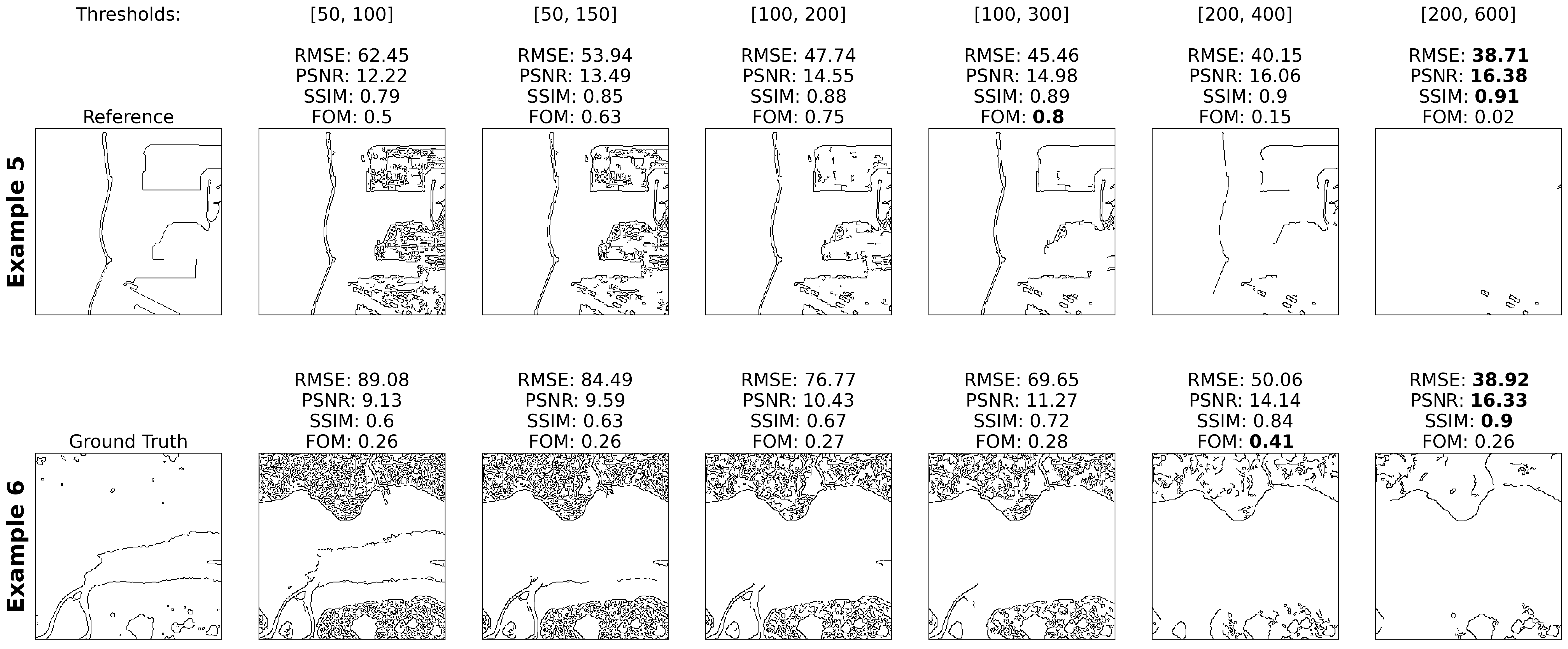}
\caption{Examples of where FOM fails to identify the best-detected edge map. The bold numbers give the best value for each metric. The NIR band is used as input for Canny.}
\label{fig:bad_examples}
\end{figure}

Figure~\ref{fig:bad_examples} also gives Example 6. Here we do not see the expected behaviour of less noisy edges being detected and the coastline edges remaining as we increase the threshold. We could consider these cases too noisy as the non-coastal edges are more prominent than coastal edges. We identified 15 of these cases. When they were removed form the analysis, we get similar results to before. That is RMSE, PSNR, SSIM and FOM could identify the best edge maps for 5.0\%, 5.0\%, 11.3\% and 73.8\% of the 80 images respectively. Additionally, FOM could identify the same or better map 93.8\% of the time. 

\psubsection{Reasons for trends}
From the above analysis, it is clear that the metrics based on MSE fail to identify the best edge map. When understanding why they fail, we should consider that comparing edge maps is a special case of comparing two images. The pixel values can only take on values of 1 or 0. This means that $(E_{i,j} - G_{i,j})^2$ will take on a value of 1 for incorrectly labelled pixels and 0 otherwise and we can reformulate MSE in terms of the confusion-matrix-based measures: 
$$MSE \propto FP +FN $$
$$RMSE \propto FP +FN $$
$$PSNR \propto \frac{1}{FP + FN} $$

Figure~\ref{fig:TP_FP_examples} shows that as we increase the thresholds fewer pixels are labelled as edges. This leads to fewer FPs and more FNs. These effects will have opposite impacts on MSE. In Figure~\ref{fig:fp_fn}, we can see that FPs tend to decrease at a higher rate than FNs increase. This leads to an improvement in MSE as we increase the thresholds. 

\begin{figure}[h]
\centering
\includegraphics[width=0.99\textwidth]{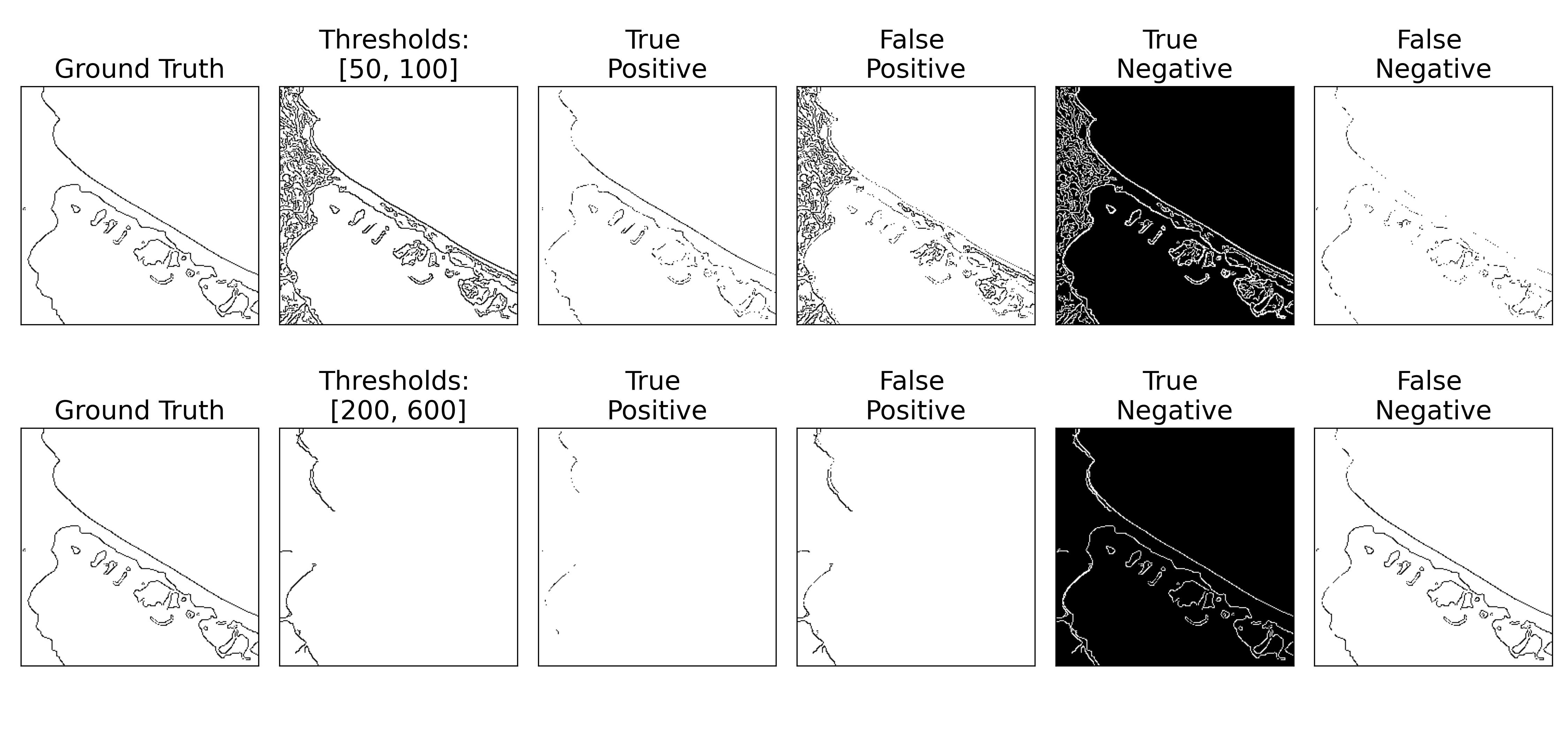}
\caption{The effect on confusion matrix measures as we increase the Hysteresis. Only the minimum and maximum thresholds are given. The NIR band is used as input for Canny.}
\label{fig:TP_FP_examples}
\end{figure}

\begin{figure}[h]
\centering
\includegraphics[width=0.99\textwidth]{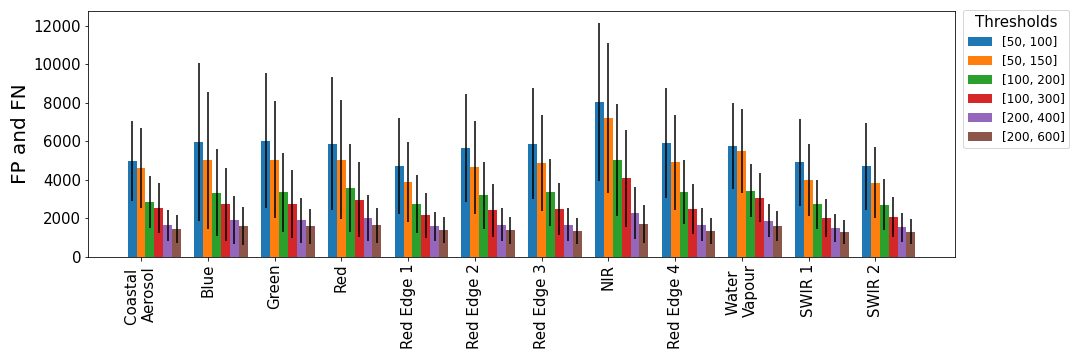}
\caption{The average sum of FP and FN for each band. The value decreases as we increase the thresholds. This indicates that FPs tend to decrease at a higher rate than FNs increase.}
\label{fig:fp_fn}
\vspace{-0.5cm}
\end{figure}

SSIM can also be reformulated in terms of confusion-matrix measures (see the appendix for proof). In this case, the relationship is not as clear but SSIM has likely been affected in a similar way to MSE. That is by the monotonic changes in the confusion matrix measures as we increase the thresholds. 
$$ SSIM(G, E) \approx  \frac{4P'P (N.TP - P'P)} {({P'}^2 + P^2)(P'(N-P') +P(N-P))} $$

These monotonic changes will not necessarily be present in all edge detection experiments. Yet, the reformulation of the MSE and SSIM suggests that, for evaluating edge detection algorithms, these metrics have no advantage over those based on confusion matrix measures. The downside to these measures is they give the same weighting to incorrectly labelled pixels regardless of their distance from ground truth edges. We can see this in Figure~\ref{fig:TP_FP_examples}, where FPs appear close to ground truth edges.In comparison, FOM incorporates the distance between pixels. This difference is potentially the reason FOM has shown to be the most effective metric for evaluating this edge detection problem. 

\psubsection{Conclusions and future work}

From our experiment, we've seen that FOM was the only reliable metric used to evaluate edge detection algorithms. Based on a visual analysis of 95 coastline satellite images, FOM could select the best detected edge map 66.3\% of the time. This is compared to RMSE, PSNR and SSIM which could select the best map 6.3\%, 6.3\%  and 11.6\% of the time respectively. Additionally, FOM could select a map that was visually better than the other metrics 92.6\% of the time. We argued that the reason for this is FOM considered the distance between pixels. In comparison, RMSE, PSNR and SSIM give equal weight to incorrectly labelled edges regardless of their distance from true edges. 

It is not clear if the results of our experiments would extend to all edge detection approaches. By varying the Hysterious threshold of the Canny edge detection algorithm, we were able to compare trends in the metrics to changes in the algorithm's sensitivity to edges. Other approaches, such as deep learning,  will not have this sensitivity mechanism. Nonetheless, we were able to reformulate RMSE, PSNR and SSIM in terms of confusion-matrix metrics. This suggests that for any edge detection evaluation, these metrics do not provide advantages over confusion-matrix metrics. Following from this, it is best to use a combination of FOM and confusion-matrix metrics to evaluate edge detection algorithms. 

By considering the distance between pixels, FOM also has a meaningful interpretation for coastline detection problems---the average distance between the detected coastline and the actual coastline. If we compare the same coastline through time, this interpretation changes. FOM now quantifies the change in the coastline. It follows that FOM could be used as a proxy for coastal erosion rates and we plan to investigate this in future research.

\ack
This publication has emanated from research conducted with the financial support of Science Foundation Ireland under Grant number 18/CRT/6183. For the purpose of Open Access, the author has applied a CC BY public copyright licence to any Author Accepted Manuscript version arising from this submission. This research was conducted with the financial support of Science Foundation Ireland under Grant Agreement No.\ 13/RC/2106\_P2 at the ADAPT SFI Research Centre at University College Dublin. The ADAPT Centre for Digital Content Technology is partially supported by the SFI Research Centres Programme (Grant 13/RC/2106\_P2) and is co-funded under the European Regional Development Fund.

\bibliographystyle{IEEEbib}
\bibliography{./ref/longforms,./ref/references}

\appendix

\section{Appendix}

\subsection{Mean Square Error}

$$ MSE(E,G) = \frac{1}{N}  \sum_{j=1}^{N} (E_{i} - G_{i})^2  $$
$$ (E_{i} - G_{i})^2 = 
\begin{cases}
    0 & \text{if } E_{i} = G_{i} \\
    1 & \text{if } E_{i} \neq  G_{i}
\end{cases} $$
Referring to the confusion matrix metrics in Table~\ref{tab:confusion_matrix}, when $E_{i} \neq  G_{i}$ we have either a false positive (FP) or false negative (FN). This means we can reformulate MSE as:  
$$ MSE(E,G) = \frac{FP+FN}{N}    $$
If follows that: 
$$ RMSE(E,G) = \sqrt{MSE(E,G)} =  \sqrt{\frac{FP+FN}{N}}$$
$$ PSNR(E,G)  = 10log_{10}\frac{255^2}{MSE(E,G)} = 10log_{10}\frac{N*255^2}{FP+FN} $$

\subsection{Structural Similarity Index}
$$\mu_E = \frac{1}{N} \sum_{i=1}^{N} E_i = \frac{P'}{N} $$
$$\mu_G = \frac{1}{N} \sum_{i=1}^{N} G_i = \frac{P}{N}$$
\begin{align*}
 \sigma_E^2 &= \frac{1}{N-1} \sum_{i=1}^{N} (E_i-\mu_E)^2 \\ 
 &= \frac{1}{N-1} \sum_{i=1}^{N} (E_i^2-2\mu_E E_i +\mu_E^2)\\ 
 &= \frac{1}{N-1} \sum_{i=1}^{N} (E_i-2\mu_E E_i +\mu_E^2) & \text{as } E_i\in \{0,1\}\\ 
 &= \frac{1}{N-1} (P' -2\mu_EP' +N\mu_E^2) \\
 &= \frac{1}{N-1} (P' -2\frac{{P'}^2}{N} +\frac{{P'}^2}{N}) & \text{as } \mu_E = \frac{P'}{N} \\
 &= \frac{P'(N-P')}{N(N-1)} 
\end{align*}
$$ \sigma_G^2  =  \frac{P(N-P)}{N(N-1)} $$ 

\begin{align*}
 \sigma_{E,G} &= \frac{1}{N-1} \sum_{i=1}^{N} (E_i-\mu_E)(G_i-\mu_G) \\ 
 &= \frac{1}{N-1} \sum_{i=1}^{N} (E_iG_i-\mu_E G_i-\mu_G E_i +\mu_E \mu_G)\\ 
 &= \frac{1}{N-1}(TP-\mu_E P-\mu_G P' +N\mu_E \mu_G)\\ 
 &= \frac{1}{N-1}(TP-\frac{P'P}{N} -\frac{P'P}{N} +\frac{P'P}{N}) \\ 
 &= \frac{N.TP-P'P}{N(N-1)}
\end{align*}

$$ SSIM(G, E) = [l(G, E)]^\alpha.[c(G, E)]^\beta.[s(G, E)]^\gamma $$
Following from~\cite{ssimpaper}, if we set $\alpha = \beta = \gamma = 1 $ and $ C_3 = C_2 /2 $ then the equation simplies to: 

\begin{align*}
SSIM(G, E) &= \frac{(2\mu_E\mu_G+C1)(2\sigma_{E,G}+C2)}{(\mu_E^2+\mu_G^2+C1)(\sigma_E^2+\sigma_G^2+C2)}\\
& \approx  \frac{(2\mu_E\mu_G)(2\sigma_{E,G})}{(\mu_E^2+\mu_G^2)(\sigma_E^2+\sigma_G^2)} \\
&= \frac{4P'P (N.TP - P'P)}{({P'}^2 + P^2)(P'(N-P') +P(N-P))}
\end{align*}

\subsection{Remove images}
The images listed below were removed from the SWED test set for this analysis. The first image  was removed as the mask is flipped---land is labelled as 1 and water 0. The other two are removed as part of the land has not been labelled as land. 

\text{S2A\_MSIL2A\_20190803T025551\_N0213\_R032\_T54XWG\_20190803T043943\_image\_0\_0.tif}

\text{S2A\_MSIL2A\_20190901T101031\_N0213\_R022\_T34VDM\_20190901T130348\_image\_0\_0.tif}

\text{S2A\_MSIL2A\_20200405T100021\_N0214\_R122\_T34VDM\_20200405T115512\_image\_0\_0.tif}

\end{paper}

\end{document}

%% file: figures/confusion_matrix.tex
\begin{table}[h]
\centering
\begin{tabular}{lllll}
                                     &                                 & \multicolumn{2}{c}{\textbf{Prediction}}                                                     &                                          \\ \cline{3-4}
                                     & \multicolumn{1}{l|}{}           & \multicolumn{1}{l|}{\textbf{1}}              & \multicolumn{1}{l|}{\textbf{0}}              &                                          \\ \cline{2-5} 
\multicolumn{1}{c|}{\textbf{Actual}} & \multicolumn{1}{l|}{\textbf{1}} & \multicolumn{1}{l|}{True Positive (TP)}      & \multicolumn{1}{l|}{False Negative (FN)}     & \multicolumn{1}{l|}{Actual Positive (P)} \\ \cline{2-5} 
\multicolumn{1}{c|}{\textbf{Value}}  & \multicolumn{1}{l|}{\textbf{0}} & \multicolumn{1}{l|}{False Positive (FP)}     & \multicolumn{1}{l|}{True Negative (TN)}      & \multicolumn{1}{l|}{Actual Negative (N)} \\ \cline{2-5} 
\multicolumn{2}{l|}{}                                                  & \multicolumn{1}{l|}{Predicted Positive (P')} & \multicolumn{1}{l|}{Predicted Negative (N')} & \multicolumn{1}{l|}{Total (T)}           \\ \cline{3-5} 
\end{tabular}
\caption{Confusion matrix for pixel classification. Water pixels are represented by a value of 1 and land pixels are represented by a value of 0. }
\label{tab:confusion_matrix}
\end{table}

%% file: figures/threshold_counts.tex
\begin{table}[h]
\begin{center}
\begin{tabular}{|l|l|l|l|l|l|}
\hline
\textbf{Threshold 1} & \textbf{Threshold 2} & \textbf{RMSE} & \textbf{PSNR} & \textbf{SSIM} & \textbf{FOM} \\ \hline
50                   & 100                  & 0             & 0             & 0             & 2            \\ \hline
50                   & 150                  & 0             & 0             & 0             & 7            \\ \hline
100                  & 200                  & 2             & 2             & 2             & 12           \\ \hline
100                  & 300                  & 1             & 1             & 3             & 30           \\ \hline
200                  & 400                  & 4             & 4             & 20            & 29           \\ \hline
200                  & 600                  & 88            & 88            & 70            & 15           \\ \hline
\end{tabular}
\end{center}
\caption{The counts of best-performing thresholds for each metric. For each metric, we take the best-performing image across the 12 spectral bands. Using RMSE, PSNR and SSIM the highest threshold would be chosen the most often. For FOM thresholds {[}100, 300{]} would be chosen the most often. This is followed closely by {[}200, 400{]}. }
\label{tab:threshold_counts}
\vspace{-0.5cm}
\end{table}